\documentclass{article}
\usepackage{spconf,amsmath,graphicx}
\usepackage[caption=false]{subfig}
\usepackage{url}
\usepackage{hyperref}
\usepackage{cite}
\usepackage{placeins}
\def\G{{\mathcal G}}
\def\W{{\mathcal W}}

\title{Spotting the Difference: Context Retrieval and Analysis for Improved Forgery Detection and Localization}
\name{Joel Brogan\(^1\), Paolo Bestagini\(^2\), Aparna Bharati\(^1\), Allan Pinto\(^{1,3}\), Daniel Moreira\(^1\)}
\secondlinename{Kevin Bowyer\(^1\), Patrick Flynn\(^1\), Anderson Rocha\(^{1,3}\), and Walter Scheirer\(^1\)}
\address{\(^1\)Department of Computer Science and Engineering, University of Notre Dame, US\\
\(^2\)Department of Electronics, Information and Bioengineering, Politecnico di Milano, Italy\\
\(^3\)Institute of Computing, University of Campinas, Brazil}

\begin{document}
\maketitle
\ninept
\begin{abstract}
As image tampering becomes ever more sophisticated and commonplace, the need for image forensics algorithms that can accurately and quickly detect forgeries grows. In this paper, we revisit the ideas of image querying and retrieval to provide clues to better localize forgeries. We propose a method to perform large-scale image forensics on the order of one million images using the help of an image search algorithm and database to gather contextual clues as to where tampering may have taken place. In this vein, we introduce five new strongly invariant image comparison methods and test their effectiveness under heavy noise, rotation, and color space changes. Lastly, we show the effectiveness of these methods compared to passive image forensics using Nimble~\cite{nist_2016}, a new, state-of-the-art dataset from the National Institute of Standards and Technology (NIST).
\end{abstract}

\begin{figure}[th!]
\begin{center}

\includegraphics[width=2.8in]{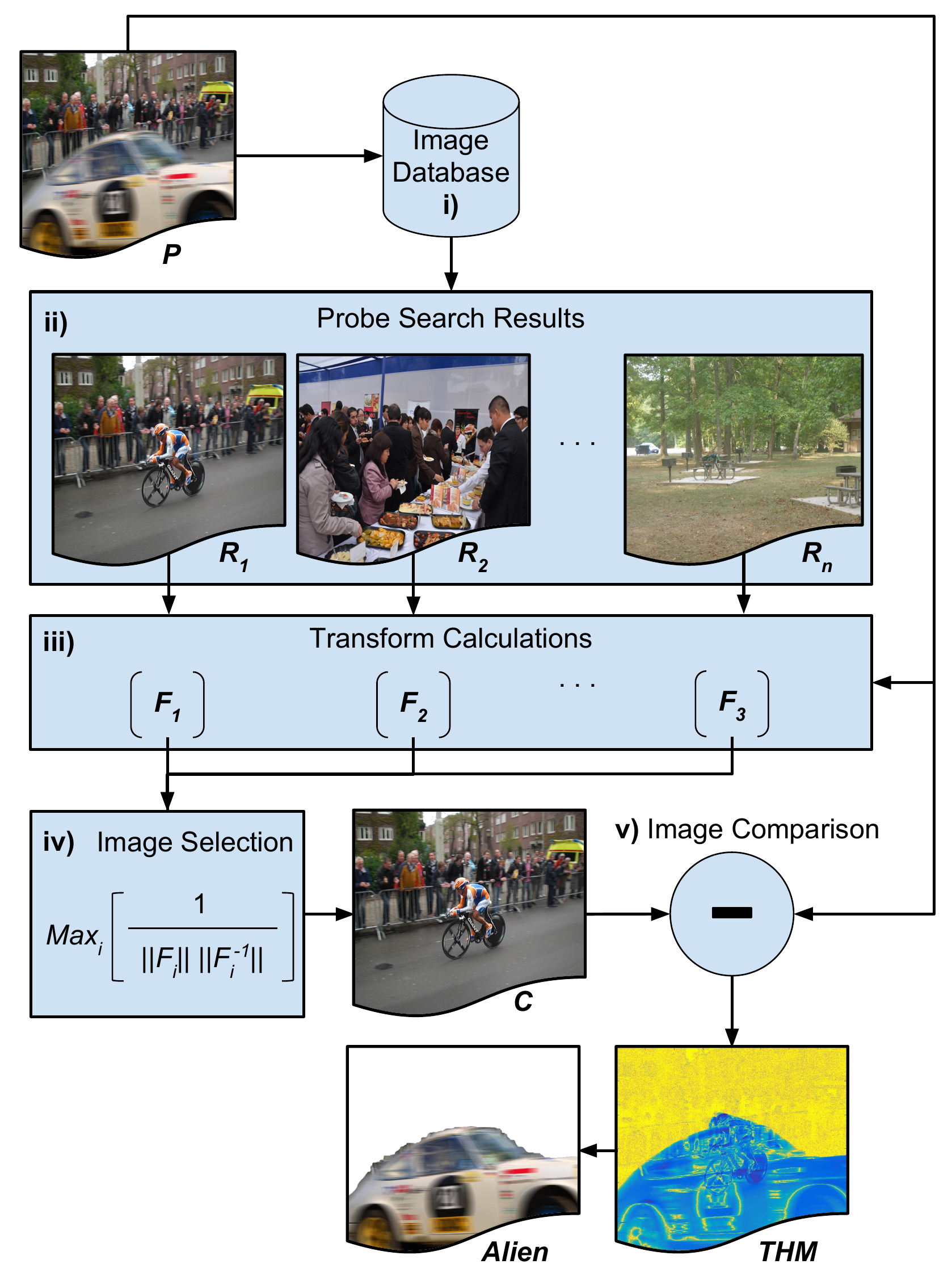}
\caption{An overview of the context-based search-and-compare framework. Probe image \textit{P} is used as input to a database search in step \textbf{i)}, which returns a list of results \textit{R}, shown in step \textbf{ii)}. The transforms between \textit{P} and \textit{R} are found in step \textbf{iii)}. The top-related image \textit{C} is chosen in step \textbf{iv)}. In step \textbf{v)}, \textit{C} is compared to \textit{P} using one of five proposed algorithms from Section~\ref{sec:comparison}, yielding a tampering heat map (THM) used to extract the alien region of \textit{P}.}
\label{fig:pipeline_overview} 
\end{center}
\vspace{-0.3in}
\end{figure}

\section{Introduction}
\label{sec:intro}
 Now that advanced and readily available photo editing software exists, image tampering has become ubiquitous, and the traces left behind by such modifications are becoming increasingly hard to detect. Regardless of intention, this trend has undermined the value of images as viable evidence in a number of domains. To examine cases of tampering, a two-fold task can be pursued. First, tampering within an image must be detected without the use of pre-embedded information (\textit{e.g.}, a key). This is known as \textbf{Passive Digital Image Forensics (PDIF)}~\cite{Rocha_2011_CSUR}. Second, the tampered area must be accurately localized if it is to be considered for further analysis.

In this paper, we improve upon the basic idea of image comparisons for contextual-clue-based PDIF offered in Gaborini et al.~\cite{gaborini2014multi}. In a PDIF scenario, a \textbf{contextual clue} can be interpreted as the incongruities between the image in question and images collected from outside sources. While~\cite{gaborini2014multi} presents a minimal method for contextual-clue-based image forensics, our work proposes a fully-automatic, efficient, and scalable  search-and-compare framework for image forensics. Additionally, our work offers highly noise-invariant comparison algorithms. This framework treats the image under question as a search probe and does not rely on having well-posed and pre-gathered images, as in \cite{gaborini2014multi}. Accordingly, the system described in this paper provides improved performance over traditional PDIF methods, and automatic extraction of alien regions within composite images. These properties make the system potentially useful for subsequent tasks like image provenance analysis, \textit{i.e.}, the study of how and when an image has been modified over time. The proposed system utilizes a fast, light-weight search engine based on KD-Trees~\cite{silpa2008optimised} optimized for SURF descriptors~\cite{bay2006surf} to retrieve a set of near-duplicate images related to the probe. These images are then compared to the probe to produce \textit{contextual} clues as to the location of modifications. An overview of the entire framework is shown in Figure~\ref{fig:pipeline_overview}. 

Three main modifications are typically performed on images: splicing, copy-move, and re-sampling. As a case study, this paper considers only instances of image splicing, also known as image compositing. In general, splicing includes an original \textbf{Host Image}, onto which regions from one or more \textbf{Donor Images} are added to create a \textbf{Composite Image}. Regions from a donor image present on a host image are known as \textbf{Alien Regions}. Most classic PDIF approaches exploit the nature of the digital photographic process to find evidence of these alien regions by using the data contained within the image in question. In contrast, our method uses outside information (\textit{i.e.}, context) gathered from a search engine to collect such evidence.

Within our search-and-compare framework, we analyze the performance of five novel signal-processing-based image comparison techniques for extracting \textbf{Tampering Heat Maps (THMs)} that indicate regions of the image where tampering might have occurred. We intentionally chose to avoid deep learning-based techniques because we found that sufficient training data that fully captures the variations seen in manipulated images is not readily available, nor is it clear how a comprehensive training dataset could be assembled. Additionally, the computational resources required for deep learning techniques are prohibitive to the massively scalable and efficient system needed to accomplish the proposed task.
Using the five signal-processing based comparison techniques, we test the proposed image-search-and-compare algorithms on the Nimble dataset, a newly developed dataset from NIST built for the task of image provenance \cite{nist_2016}, mixed in with a set of 1 million distractor images. Finally, we compare the results of our  image search method to a set of 13 state-of-the-art PDIF splicing algorithms to show the effectiveness of the proposed method.
% \begin{figure}[tb]
% \begin{center}

% \includegraphics[width=2.5in]{fig_putinphotoshop.pdf}
% \caption{The \textbf{composite image} c) from slate.com is created by \textbf{donor image} a) and \textbf{host image} b). The subsequent tree created denotes the \textbf{image provenance}. The segmented splice area of the composite is shown in d)}
% \label{fig:trump_putin_provenance}
% \end{center}
% %\vspace{-0.2in}
% \end{figure}

\section{Related Work}
\label{sec:relatedwork}

Forgery detection and localization based on single image analysis can be performed using a variety of traditional PDIF methods \cite{Cozzolino2014, zampAcc}. For example, one set of techniques exploit JPEG artifacts~\cite{lin2009fast,bianchi2011improved,amerini2014splicing,li2009passive,ye2007detecting,bianchi2012image,farid2009exposing}. Another set of approaches utilize Color Filter Array (CFA) footprints \cite{ferrara2012image,dirik2009image}. Yet another set of methods deal with detecting natural noise inconsistencies within spliced images \cite{mahdian2009using,lyu2014exposing,wagner_2015}. Additionally, methods based on Error Level Analysis (ELA) \cite{krawetz2007pictures} can be used.% to detect image splicing.
%A large scale study of many of these algorithms has been performed in~\cite{zampAcc}.

More recently, image provenance research has shown the possibility of conducting even deeper forensic analysis by jointly considering sets of correlated images \cite{de2010exploring,dias2010first,dias2012image}.
Indeed, if multiple images are available, it is possible to achieve robust forgery localization results through image comparison \cite{Bestagini2016}. For instance, using scaled thumbnail meta-data of images to localize forgeries~\cite{kirchner2013impeding} can provide high-accuracy localization maps of image tampering. When thumbnail data is unavailable, near-duplicate images have also been used to build THMs from contextual clues as to where forgeries occur~\cite{gaborini2014multi}. While these ideas show promise, they provide no automatic method of retrieving contextually relevant images for comparison. Additionally, the comparison methods found in the literature offer relatively poor invariance to color changes, noise, morphing, and compression between images.

By always considering a scenario with multiple images, sophisticated methods for patch comparison using deep learning have been proposed~\cite{zagoruyko2015learning,han2015matchnet}. However, the corresponding models were trained for highly specific keypoint matching scenarios. Techniques like these do not capture the variations present in  realistic forgeries, and thus cannot be used as-is in a real-world forensic scenario.

%Subsequently, relatively simple comparison techniques have been used  in multiple works to construct dependency and provenance graphs~\cite{de2010exploring,dias2010first,dias2012image}, however these methods assume that the images in question are already obtained and grouped together. 

%In this paper, we propose a fully automatic system for %image search, retrieval, and comparison for the purpose %of image forensics and provenance analysis. Indeed, to %the best of our knowledge, no systems exist to fully %solve this problem.

\section{Image Search Algorithm}
\label{sec:search}
The first step of the search-and-compare process is image search, as shown in Fig.~\ref{fig:pipeline_overview}. The search engine must adhere to multiple constraints. First, the system must provide fast and scalable indexing and searching. For the proposed search-and-compare method to be effective, we must have an extremely large database (on the order of one million images in this work) to compare against. Second, the system must be sensitive to near-duplicate and contextually similar entries, and tolerant of perceptually similar entries. Specifically, the search engine should return only images that provide useful comparisons with the probe in question.

In our proposed search method, we extract 500 SURF keypoints \cite{bay2006surf} with the relative 64-dimensional descriptors to describe each image. We utilize a KD-tree forest scheme similar to what was used in~\cite{silpa2008optimised} to scalably index and search the descriptors in the database. Using SURF features helps to ensure that image representations are clustered by context of local objects within the image, as opposed to an auto-encoder approach that may cluster by the global appearance of an image. This methodology provides a higher likelihood of returning images that contain objects that can be directly compared to the probe~\cite{krizhevsky2011using}. According to the described scheme, once a probe image $P$ is queried, the system returns a set of $N$ contextually similar images along with any possible near-duplicate images $R_{n}, n \in [1, N]$.
	
\section{Image Comparison Framework}
\label{sec:comparison}
%\textbf{Image Filtering}
Once the images $R_{n},n \in [1, N]$ are retrieved from the image database using probe image $P$ as a query, they must be sorted and filtered to ensure only truly relevant images to our probe are compared. To accomplish this goal, for each image $R_{n}$, SURF keypoints and features are re-calculated. Then, the $3 \times 3$ affine matrix $F_i$ mapping points of $R_{n}$ to the coordinate system of $P$ is computed using keypoint matching and the MSAC method, allowing for tighter geometric constraints than RANSAC \cite{choi1997performance}. To generate a list of images with content that best geometrically matches the probe, we rank each \(R_{n}\) by the Reciprocal Frobenius Condition of its linked affine transform $F_n$ as
\begin{equation}
\text{RFN}_{n}=\frac{1}{\| F_{n}  \|\| F_{n}^{-1} \| },
\end{equation}
where $\| \cdot \|$ is the Frobenius norm of a matrix.

We assert that the greater the RFN value, the more suitable $R_n$ will be to the comparison task. Therefore, even though multiple images could be used to provide multiple clues, we decided in this work to only select as comparison image \(C\) the image $R_n$ with the highest RFN Value warped using the affinity matrix $F_n$
\begin{equation}
    C = \text{warp}(R_{\hat{n}}, F_{\hat{n}}), \quad \hat{n} = \arg\max_{n}(\text{RFN}_n),
\end{equation}
where $\text{warp}$ applies the affine transform to the image.
Once this image has been selected, we must compare the probe image \(P\) to the result image \(C\) to produce a THM as shown in Fig.~\ref{fig:pipeline_overview}. To achieve a reliable comparison, an algorithm must overcome differences in image noise, colorspace changes, and slight rotations and translation. For this purpose, we propose he following five algorithms.

\textbf{1. PSNR of Gaussian Image Residual. (IRPSNR)} We define \(\G(I,\sigma_\G)\) to be the convolution of image \(I\) with a Gaussian kernel with standard deviation \(\sigma_\G\). We set $\sigma_\G=4$, as we found it to provide optimal local blurring to allow for invariance to small translations and rotations. To generate a tampering heat map, we compute the pixel-wise Peak Signal to Noise Ratio (PSNR) between the Gaussian blurred versions of \(P\) and \(C\) as
\begin{equation}
    \text{THM}_\text{PSNR} = \log_{10}{\frac{1}{\left | \G(P,\sigma_\G)-\G(C,\sigma_\G) \right |^{2}+1}},
\end{equation}
where all operations are pixel-wise, and the plus $1$ in the denominator is used for regularization. Portions of $P$ matching the respective portions of $C$ will contribute to the $\text{THM}_\text{PSNR}$ with high values. Tampered areas should be exposed by low $\text{THM}_\text{PSNR}$ values.

\textbf{2. Pseudo-PRNU Patch-wise Comparison.} Images shot with the same camera are characterized by a multiplicative noise pattern known as Photo-Response Non-Uniformity (PRNU) \cite{Lukas2006}. This noise residual is characteristic of the capturing device, and can be used for camera attribution \cite{Lukas2006}, for tampering localization \cite{Chen2008}, or even to assess whether two images come from the same device \cite{Goljan2006}. As near-duplicate images are acquired by the same camera by definition, we can rely on image noise patch-wise comparison to detect local inconsistencies due to splicing.

Given two corresponding patches of $P$ and $C$ (\textit{e.g.}, the first $64\times64$ pixel block in the top-left corner of each image), PRNU information extracted from those patches should correlate very well if only global transformations (\textit{e.g.}, color corrections, blurring, compression, etc.) have been applied to $P$ or $C$. Conversely, PRNU information does not correlate at all if one of the two patches has been spliced from a picture obtained from a different device.

To exploit this property, let us define the noise residuals 
\begin{equation}
	\tilde{C} = C - \W(C), \qquad	\tilde{R} = R - \W(R),
\end{equation}
where $\W(\cdot)$ is the wavelet-based denoising operation used in~\cite{Lukas2006}. According to \cite{Lukas2006}, the computed $\tilde{C}$ and $\tilde{R}$ contain PRNU traces. Therefore, it is possible to correlate them patch-wise for the THM
\begin{equation}
	\text{THM}_\text{noise} = \text{average}\left(\frac{\tilde{C} \cdot \tilde{R}}{||\tilde{C}|| \cdot ||\tilde{R}||}\right),
\end{equation}
where `$\cdot$' represents pixel-wise multiplication, $||\cdot||$ returns the Frobenius norm, and $\text{average}(\cdot)$ computes the moving average on $64 \times 64$ pixel blocks. The mask $\text{THM}_\text{noise}$ should present high values corresponding to areas that are common to $P$ and $C$, and low values in tampered regions.

\textbf{3. Structural Similarity Comparison. (SSIM)}
This method uses the calculated pixel-wise Structural Similarity Index Measure (SSIM) between images \(P\) and \(C\).  We define the structural similarity-based THM as 
\begin{align}
&A = (2\mu _{P}\mu_{C}+(0.01 D)^{2})(2\sigma_{PC}+(0.03 D)^{2}),\\
&B = (\mu_{P}^{2}+\mu_{C}^{2}+(0.01 D)^{2})(\sigma_{P}^{2}+\sigma_{C}^{2}+(0.03 D)^2),\\
&\text{THM}_\text{SSIM} = \frac{A}{B},
\end{align}
where \(\mu\) and \(\sigma\) are the local neighborhood means and standard deviations of \(P\) and \(C\) with a neighborhood radius of 32 pixels, \(\sigma_{PC}\) is the local covariance of the local image patches, and \(D\) is the dynamic contrast of the images. Similar to the PRNU-based mask, $\text{THM}_\text{SSIM}$ should assume low values in correspondence of tampered areas.

\textbf{4. HSV Histogram Patch-wise Comparison.}
From images \(P\) and \(C\), local histogram patches \(H_{P_{xy}}\) and \(H_{C_{xy}}\) are calculated using a local neighborhood radius of 13 pixels. We use the probability of the two-sample Kolmogorov-Smirnov Test~\cite{lilliefors1967kolmogorov} being equal to or more extreme than the observed value of the null hypothesis that \(\text{CDF}_{H_{P}} = \text{CDF}_{H_{C}}\), where \(\text{CDF}_{H}\) is the Cumulative Distribution Function of a given histogram patch. This value is calculated for each corresponding patch to generate a THM:
\begin{equation}
    Pr_{xy}(\max_{a}(| \mathcal{Q}(a,H_{C_{xy}})-\mathcal{Q}(a,H_{P_{xy}}) |) \; | \; H_{P_{xy}}),
\end{equation}
\begin{equation}
    \text{THM}_\text{HPC} = Pr,
\end{equation}
where \(\mathcal{Q}(a,\vec{b})\) is the proportion of \(\vec{b}\) less than or equal to \(a\). This allows us to test, in a manner invariant to small rotations, the idea that each patch contains a similar distribution.

\textbf{5. PatchMatch 2.1.}
For an additional method, we use the PatchMatch algorithm \cite{barnes2009patchmatch} for image comparison. Specifically, we utilize the rotation and scale-invariant version, using 20 iterations. To generate the relative $\text{THM}_\text{PM}$, we calculate the L2 match distance of patches within the image. In other words, we associate to each $8\times8$ patch of $P$ the L2 distance from the patch of $C$ that best approximates it. If a patch of $P$ cannot be well approximated with any patch of $C$ (\textit{i.e.}, in case of tampering), its L2 distance will be high.

\begin{figure}[t]
\begin{center}
\includegraphics[width=3in]{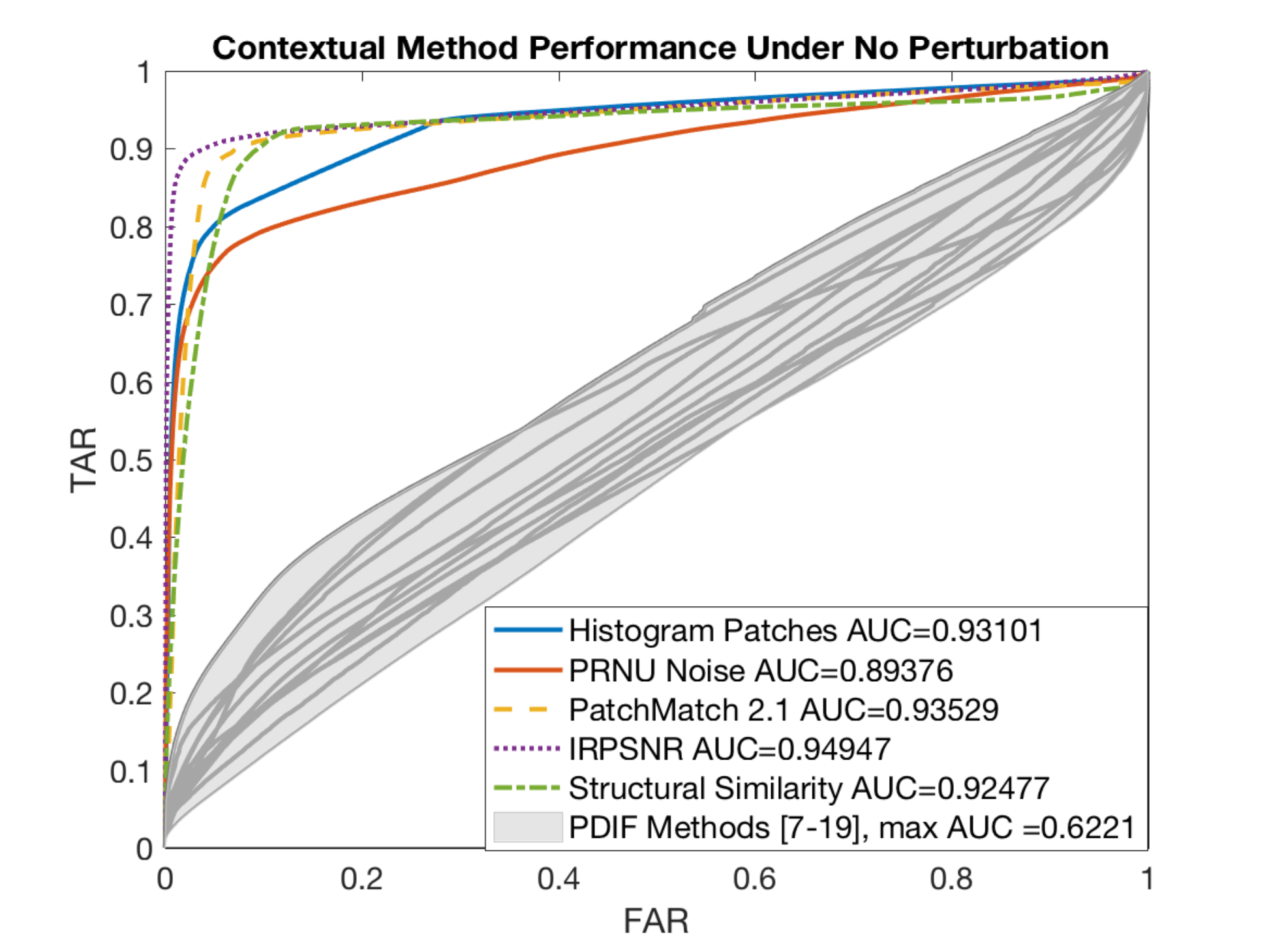}
\caption{Five variants of the proposed context-based search-and-compare framework compared against 13 widely-used PDIF techniques. The gray area represents the maximum and minimum performance of each PDIF algorithm. Even the worst performing contextual method, PRNU matching, performs \textbf{44\%} better than the top PDIF comparison algorithm. For the best performing variants, we see that IRPSNR performs \textbf{1.5\%} better than PatchMatch 2.1.}
\label{fig:ROCnone}
\end{center}
\vspace{-0.2in}
\end{figure}

\section{Experiments and Results}
\label{sec:Experiments}

%\begin{figure*}[t]
%\centering
%\subfloat[]{\includegraphics[width=.66\columnwidth]{fig_colorROC}\label{fig:ROCcolor}}\hfil
%\subfloat[]{\includegraphics[width=.66\columnwidth]{fig_noiseROC}\label{fig:ROCnoise}}\hfil
%\subfloat[]{\includegraphics[width=.66\columnwidth]{fig_rotationROC}\label{fig:ROCrotate}}\hfil
%\caption{}
%\end{figure*}

\textbf{Dataset.} For the purpose of our experiments, we utilize a new, state-of-the-art dataset from NIST called Nimble \cite{nist_2016}.  The dataset contains a subset specifically for splicing operations, dubbed Nimble-Splice, which contains a total of 288 probe images, each having been hand-composited from a gallery of 874 images. A host, donor, and binary tamper mask image is provided with each probe. The masks represent ground-truth data to compare our generated THMs against.

To simulate a real-world scale, we take the 874 gallery images from Nimble-Splice and add one million distractor images provided by RankOne Inc.\footnote{\url{http://medifor.rankone.io/}}. This allows us to test the effectiveness of indexing and subsequently finding relevant images for comparison. We call this hybrid dataset ``Nimble World" (NW).

\textbf{Framework Setup.} Using the method described in Section~\ref{sec:search}, we
extract features and index all 1,000,874 images into a KD-tree forest. For each probe we return the top $I=100$ results from the KD-forest search. We find that our search algorithm returns relevant results with \textit{Recall at Rank 25} = 99.5\%. The set of results is then sorted and the top scoring image is registered to the probe using the affine transform described in Section \ref{sec:comparison}. We then test all five proposed algorithms to compare the probe to the result and generate a THM. Using a sliding threshold approach to localize forgeries (i.e., image differences), we generate a ROC curve of generated THMs compared to ground-truth binary masks.% As a baseline comparison, we use the PatchMatch algorithm~\cite{barnes2009patchmatch}.

\textbf{Experiment 1.} The first experiment studies the performance of the methods we propose in this paper and 13 current state-of-the-art PDIF splicing-detection algorithms. These algorithms include JPEG artifact analysis~\cite{lin2009fast,bianchi2011improved,amerini2014splicing,li2009passive,ye2007detecting,bianchi2012image,farid2009exposing}, CFA analysis~\cite{ferrara2012image,dirik2009image}, image noise analysis~\cite{mahdian2009using,lyu2014exposing,wagner_2015}, and ELA~\cite{krawetz2007pictures}. The THMs generated by these methods were compared to the ground-truth masks using the same thresholding method to generate ROC curves. In Fig.~\ref{fig:ROCnone} we see a large performance gap between our five proposed algorithms within the search-and-compare framework, which perform the best, and the set of 13 PDIF methods from the literature.

\textbf{Experiment 2.} The second experiment analyzes the performance of each of the five proposed image-to-image comparison algorithms in the presence of non-negligible noise, color, and rotation perturbations. These perturbations, performed on the result images \(R\) that are used in individual comparisons, simulate real-world  artifacts likely to be found in images indexed from the web.

\begin{figure}[t]
\begin{center}
\includegraphics[width=3in]{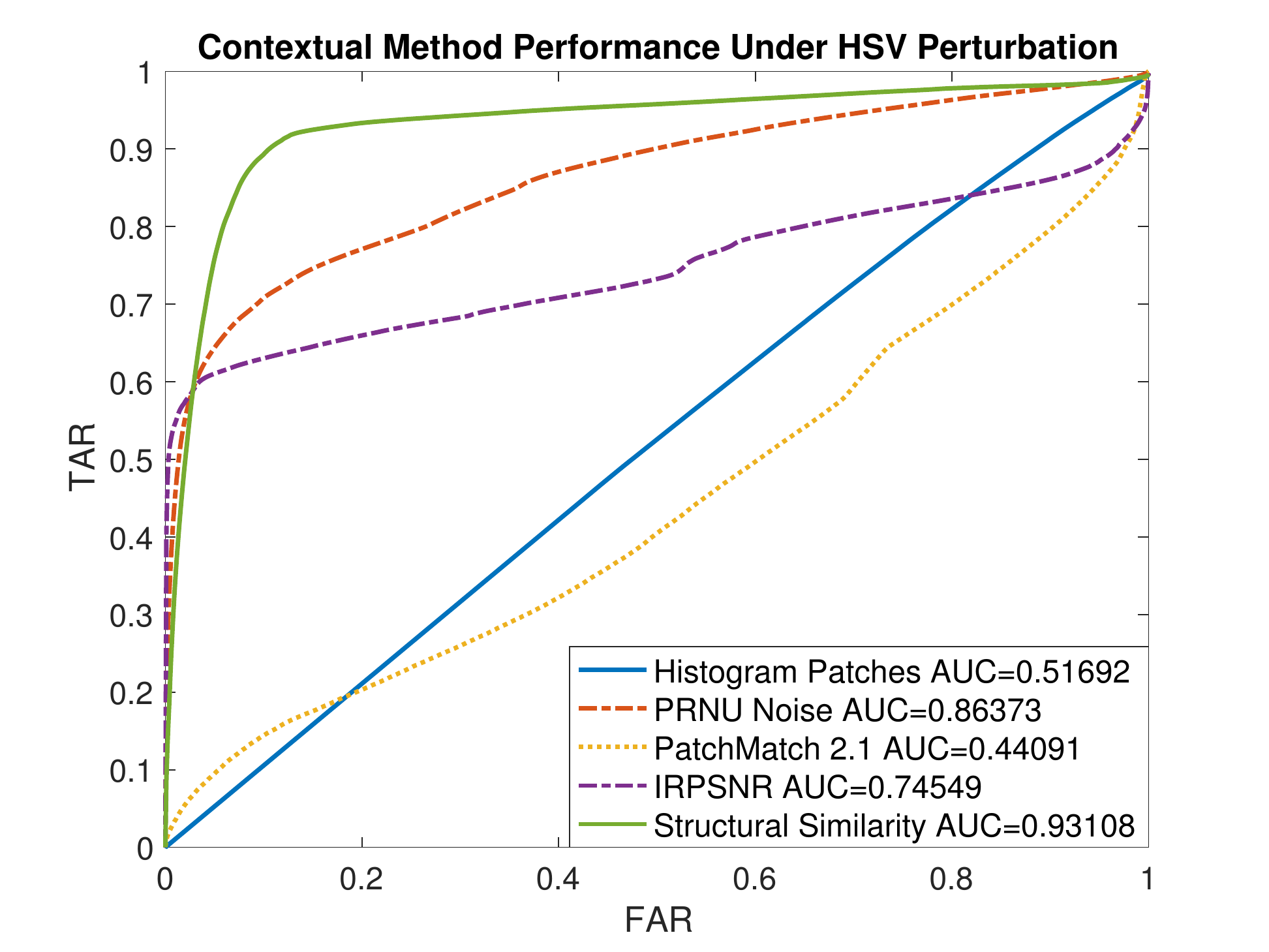}
\vskip -.5em
\caption{Performance of all five contextual image forensics methods under random HSV space transformations. The SSIM approach is most invariant to color changes, while most other algorithms suffer. Patchmatch 2.1 performs poorly in such scenarios.}
\label{fig:ROCcolor}
\vspace{-0.2in}
\end{center}
\end{figure}
\begin{figure}[t]
\begin{center}
\includegraphics[width=3in]{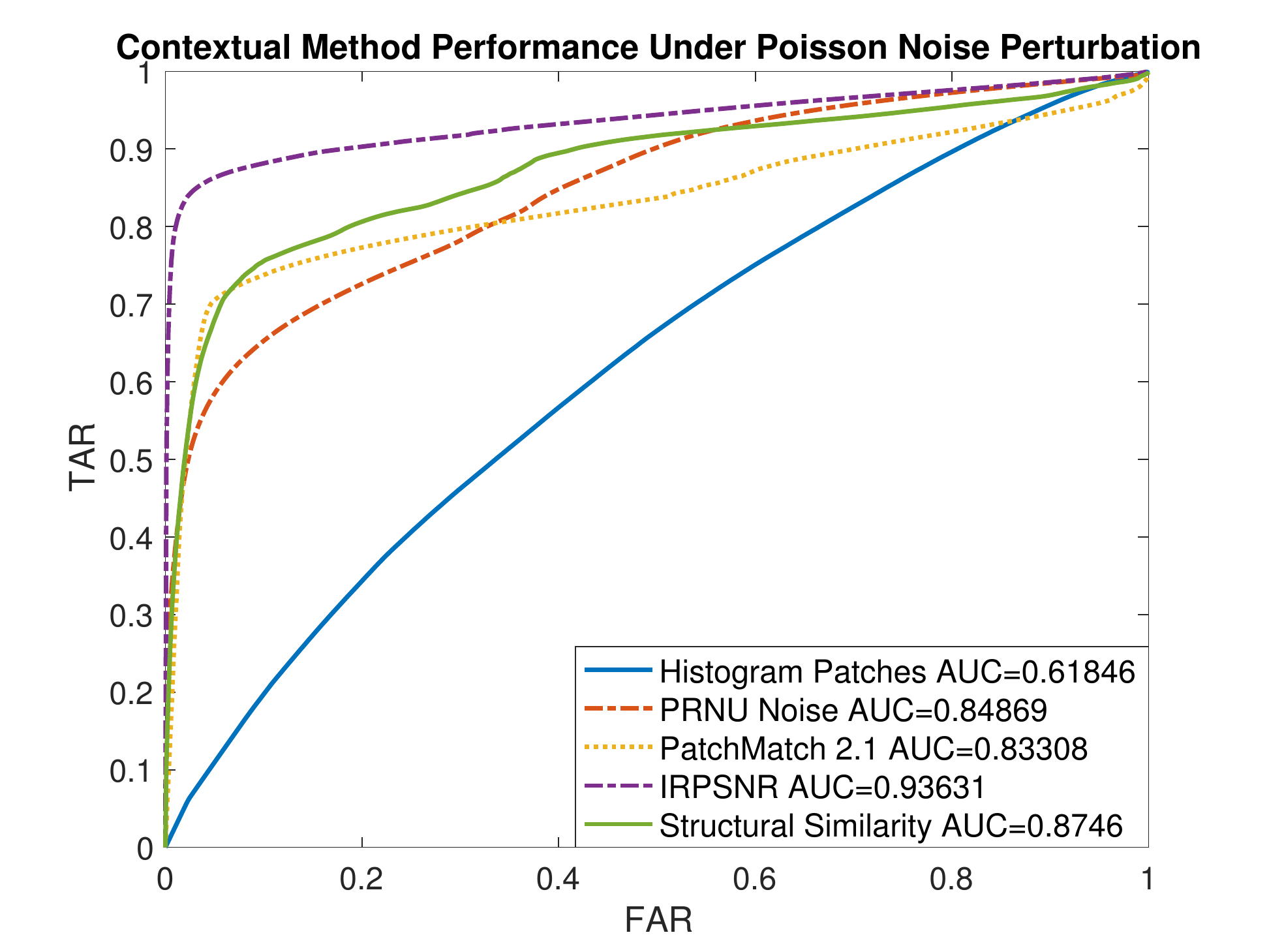}
\vskip -.5em
\caption{Performance of all five contextual image forensics methods under the addition of Poisson Noise. The IRPSNR approach is most invariant to noise.}
\label{fig:ROCnoise}
\vspace{-0.2in}
\end{center}
\end{figure}
\begin{figure}[t]
\begin{center}
\includegraphics[width=3in]{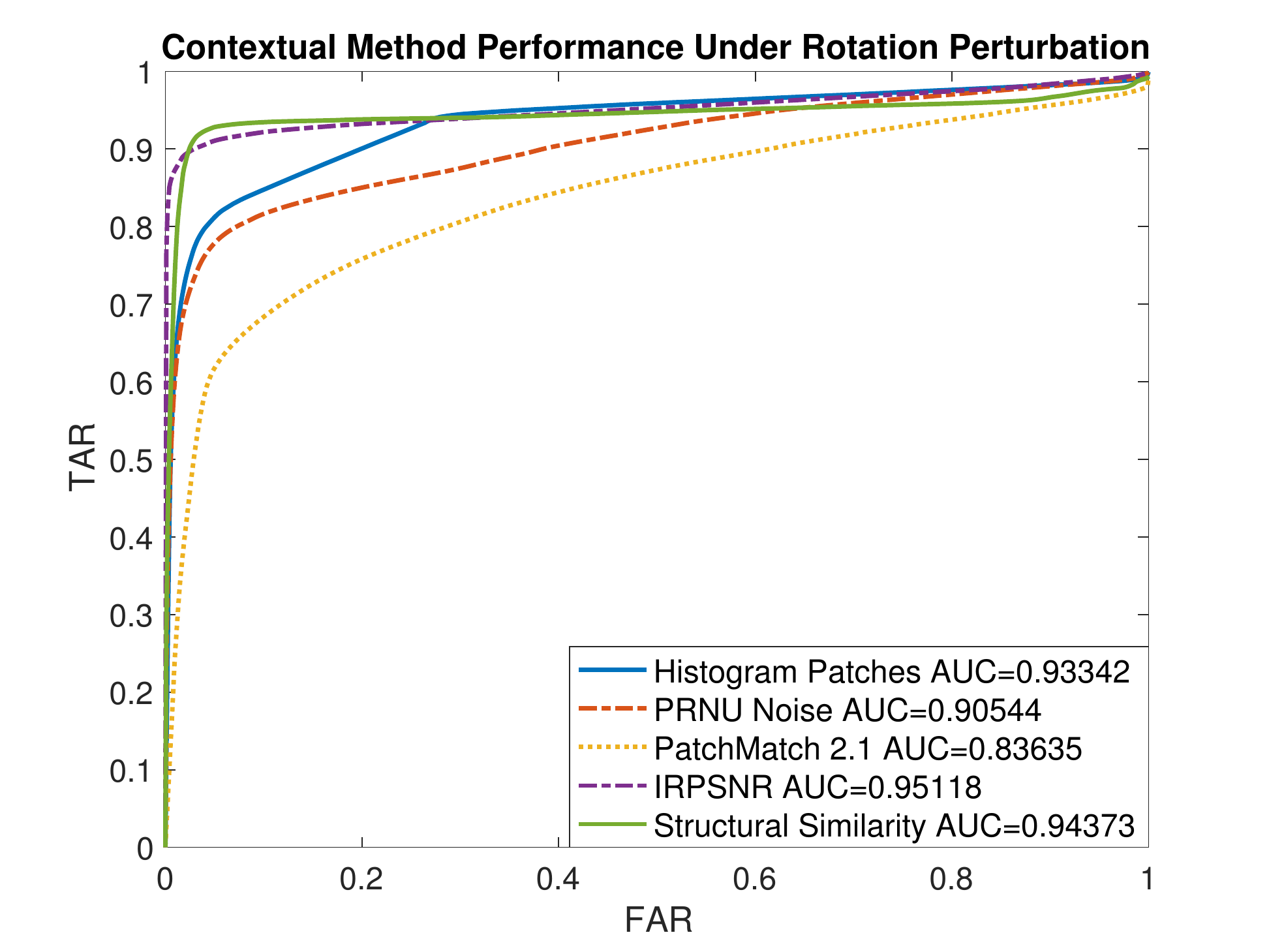}
\vskip -.5em
\caption{Performance of all five contextual image forensics methods under random small-angle image rotations. The PatchMatch approach is most negatively impacted; other algorithms perform well.}
\label{fig:ROCrotate}
\vspace{-0.2in}
\end{center}
\end{figure}

To perturb the color space of gallery images in the NW dataset, we randomly fluctuated the HSV channels of each image independently between 0 and 20\%. The results for color space perturbing can be seen in Fig.~\ref{fig:ROCcolor}. Similarly, to perturb the noise within gallery images, random amounts of Poisson noise were added to each gallery image. Results for noise can be seen in Fig.~\ref{fig:ROCnoise}. Lastly, we used random rotations between \(-15^{\circ}\) and \(15^{\circ}\) \textit{after} the result registration phase, to simulate an incorrect registration caused by erroneous keypoint matches. Results for rotation can be seen in Fig.~\ref{fig:ROCrotate}.

We concluded that a test on scale-based perturbations was not necessary after observing that algorithm performance was \textit{nearly identical} even for large-scale fluctuations. 

\section{Conclusion}
\label{sec:conclusion}

Assuming a search process that provides relevant results in the presence of a large number of distractor images, a context-based search-and-compare framework for image forensics is greatly superior at localizing areas of tampering than traditional PDIF methods.  Further, we conclude that of the methods tested, the IRPSNR method  provided the most invariance to rotation and noise space perturbations, while the SSIM method had the least performance deterioration under color space perturbation. The PRNU patch-wise comparison algorithm was the most stable over all three perturbation cases, while Histogram patch-wise comparison and PatchMatch had the least invariance to all perturbation methods. 
%From these results, we can see that a fusion of IRPSNR and SSIM methods could be employed to provide more robust THMs.

It should be noted that while the search-and-compare paradigm provides improved results over traditional PDIF methods, search-and-compare fails in cases where original or near-duplicate images are not present.  However, these instances can be detected simply by testing the maximum Reciprocal Condition of each probe-to-result transform, and thresholding at an empirically determined level.

With respect to further directions for this work, the search-and-compare paradigm introduced in this paper lends itself nicely to the task of image provenance analysis and multimedia phylogeny~\cite{dias2012image,dias2010first}. To construct accurate provenance graphs that express relationships between tampered images, we must dig down into the localized tampered objects within a composite to further determine each object's origin. The THMs produced by our framework can be easily segmented into tamper regions. These regions can then be directly analyzed to determine the nature of the tampering. 
%If the tampered region contains an alien object, that %object can be re-submitted to the system to determine the %donor image it originated from. 
Thus, the work we have described in this paper should not be treated as a standalone contribution, but placed in the wider context of digital forensics.
% Below is an example of how to insert images. Delete the ``\vspace'' line,
% uncomment the preceding line ``\centerline...'' and replace ``imageX.ps''
% with a suitable PostScript file name.
% -------------------------------------------------------------------------
% \begin{figure}[htb]

% \begin{minipage}[b]{1.0\linewidth}
%   \centering
%   \centerline{\includegraphics[width=8.5cm]{image1}}
% %  \vspace{2.0cm}
%   \centerline{(a) Result 1}\medskip
% \end{minipage}
% %
% \begin{minipage}[b]{.48\linewidth}
%   \centering
%   \centerline{\includegraphics[width=4.0cm]{image3}}
% %  \vspace{1.5cm}
%   \centerline{(b) Results 3}\medskip
% \end{minipage}
% \hfill
% \begin{minipage}[b]{0.48\linewidth}
%   \centering
%   \centerline{\includegraphics[width=4.0cm]{image4}}
% %  \vspace{1.5cm}
%   \centerline{(c) Result 4}\medskip
% \end{minipage}
% %
% \caption{Example of placing a figure with experimental results.}
% \label{fig:res}
% %
% \end{figure}

%\vfill
%\pagebreak

\bibliographystyle{IEEEbib}
\bibliography{references}

\end{document}